\documentclass[letterpaper]{article}
\usepackage{flairs}
\usepackage{times}
\usepackage{helvet}
\usepackage{courier}
\usepackage{graphicx}
\usepackage{adjustbox}
\usepackage{multirow, multicol}
\usepackage{amsmath}
\usepackage{amssymb}
\usepackage{xcolor}
\usepackage{balance}
\usepackage{hyperref}
\hypersetup{
    colorlinks,
    linkcolor={red!50!black},
    citecolor={blue!50!black},
    urlcolor={blue!80!black}
    }

\newcommand{\oursys}{$\mathsf{ZS}\mbox{-}\mathsf{ToD}$}

\newcommand{\stitle}[1]{\noindent\textup{\textbf{#1}}}
\newcommand{\smalls} {s}
\newcommand{\smalln} {n}
\newcommand{\smalli} {i}
\newcommand{\sgdx}{SGD-X}

\newcommand{\tod}{ToD}
\newcommand{\etoe}{end-to-end}
\newcommand{\zs}{Zero-Shot}
\newcommand{\celoss}{cross-entropy loss}
\title{Zero-Shot Generalizable End-to-End Task-Oriented Dialog System using Context Summarization and Domain Schema}
\author{Adib Mosharrof~\textsuperscript{\rm 1}, M.H. Maqbool~\textsuperscript{\rm 2}, A.B. Siddique~\textsuperscript{\rm 1}\\
adib.mosharrof@uky.edu,  hasanmaqbool@knights.ucf.edu, siddique@cs.uky.edu\\
    \textsuperscript{\rm 1}University of Kentucky,
    \textsuperscript{\rm 2}University of Central Florida\\  
}

\begin{document}
\maketitle
\begin{abstract}
    \begin{quote}

Task-oriented dialog systems empower users to accomplish their goals by facilitating intuitive and expressive natural language interactions.
 State-of-the-art approaches in task-oriented dialog systems formulate the problem as a conditional sequence generation task and fine-tune pre-trained causal language models in the supervised setting. This requires labeled training data for each new domain or task, and acquiring such data is prohibitively laborious and expensive, thus making it a bottleneck for scaling systems to a wide range of domains. 
To overcome this challenge, we introduce a novel \textbf{Z}ero-\textbf{S}hot generalizable end-to-end \textbf{T}ask-\textbf{o}riented \textbf{D}ialog system, {\oursys}, that leverages domain schemas to allow for robust generalization to unseen domains and exploits effective summarization of the dialog history. 
We employ GPT-2 as a backbone model and introduce a two-step training process where the goal of the first step is to learn the general structure of the dialog data and the second step optimizes the response generation as well as intermediate outputs, such as dialog state and system actions.
As opposed to state-of-the-art systems that are trained to fulfill certain intents in the given domains and memorize task-specific conversational patterns, {\oursys} learns generic task-completion skills by comprehending domain semantics via domain schemas and generalizing to unseen domains seamlessly.
We conduct an extensive experimental evaluation on SGD and SGD-X datasets that span up to 20 unique domains and {\oursys} outperforms state-of-the-art systems on key metrics, with an improvement of \textbf{+17\% on joint goal accuracy} and \textbf{+5 on inform}. 
Additionally, we present a detailed ablation study to demonstrate the effectiveness of the proposed components and training mechanism.

    \end{quote}
\end{abstract}

\section{Introductions}
Task-Oriented Dialog (ToD) systems facilitate users to achieve their goals through the utilization of human-like language interactions.
Traditionally, ToD systems have employed diverse modular architectures~\cite{molich1990improving}, incorporating separate components for Natural Language Understanding (NLU)~\cite{mairesse2009spoken,lee2019convlab}, Dialogue State Tracking (DST)~\cite{ren2018towards,lee2013structured}, Dialogue Policy (POL)~\cite{peng2018deep,le2021predictable,le2021generating}, and Natural Language Generation (NLG)~\cite{wen2015semantically,peng2020few} that are connected in a pipeline. 
Other variations of the pipeline also exist where NLU and DST are merged into a single module, named Word-DST~\cite{ramadan2018large}, while POL and NLG are integrated into a single module, called Word-POL~\cite{chen2019semantically,budzianowski2018towards}.
Moreover, End-to-End (E2E) systems have emerged, producing a natural language response directly from the user's input without using intermediate stages~\cite{bordes2016learning}. 
E2E ToD systems that include intermediate outputs (e.g., DST, POL) have gained increasing popularity recently since such systems can utilize intermediate outputs to facilitate effective communication with external APIs~\cite{zhang2020task}.

\begin{figure}[!t]
    \centering
    \includegraphics[width=\linewidth]{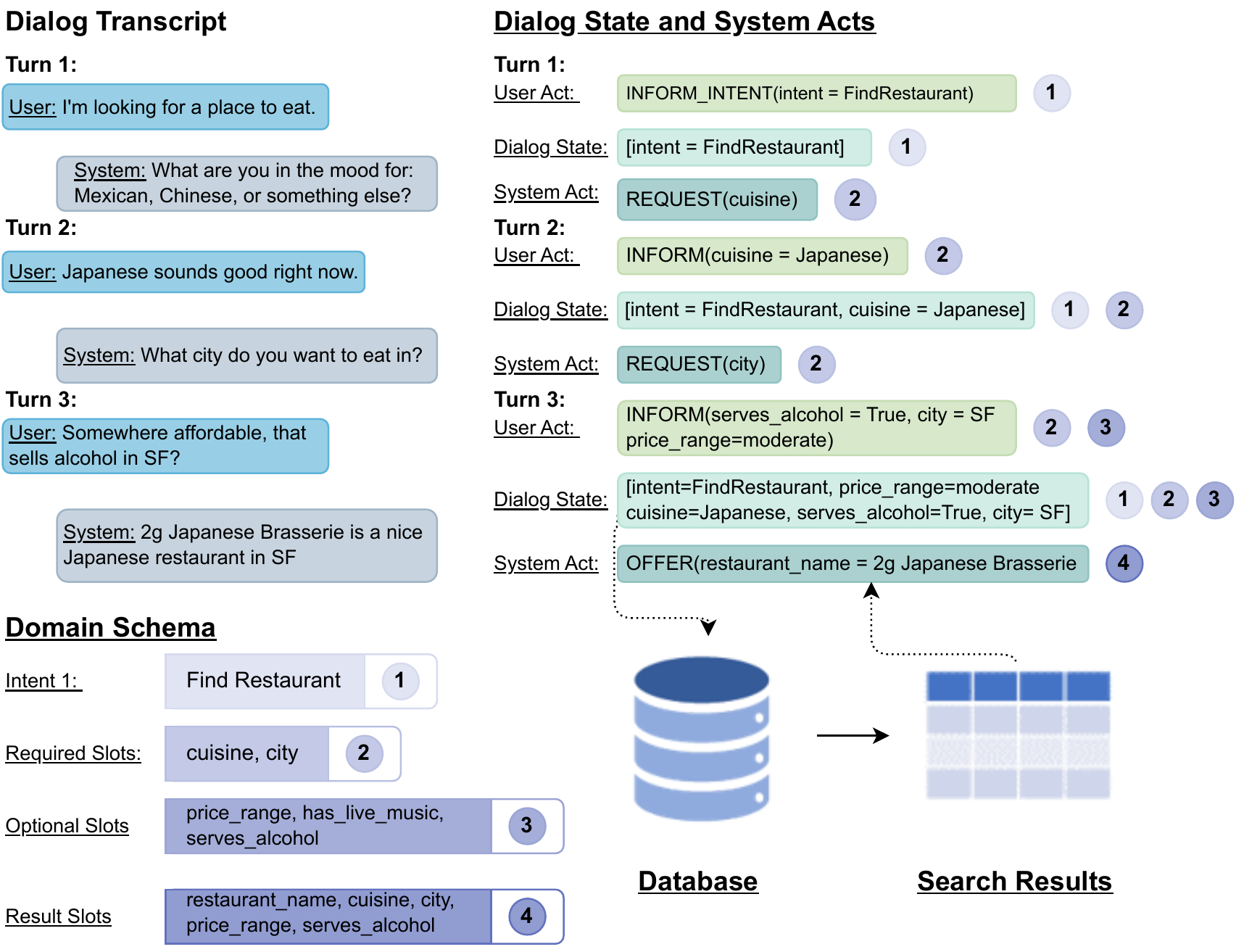}
    \vspace{-10pt}
    \caption{
Overview of {\oursys}: the domain schema facilitates estimating dialog state, system actions, and system response irrespective of whether the model was trained on that domain or not.
Parts of the schema that assist in the generation are grouped by similar colors.
    }
    
    \label{fig:approach}
    \vspace{-12pt}
\end{figure}

\begin{figure*}[t]
   \centering
   \includegraphics[width=0.95\linewidth]{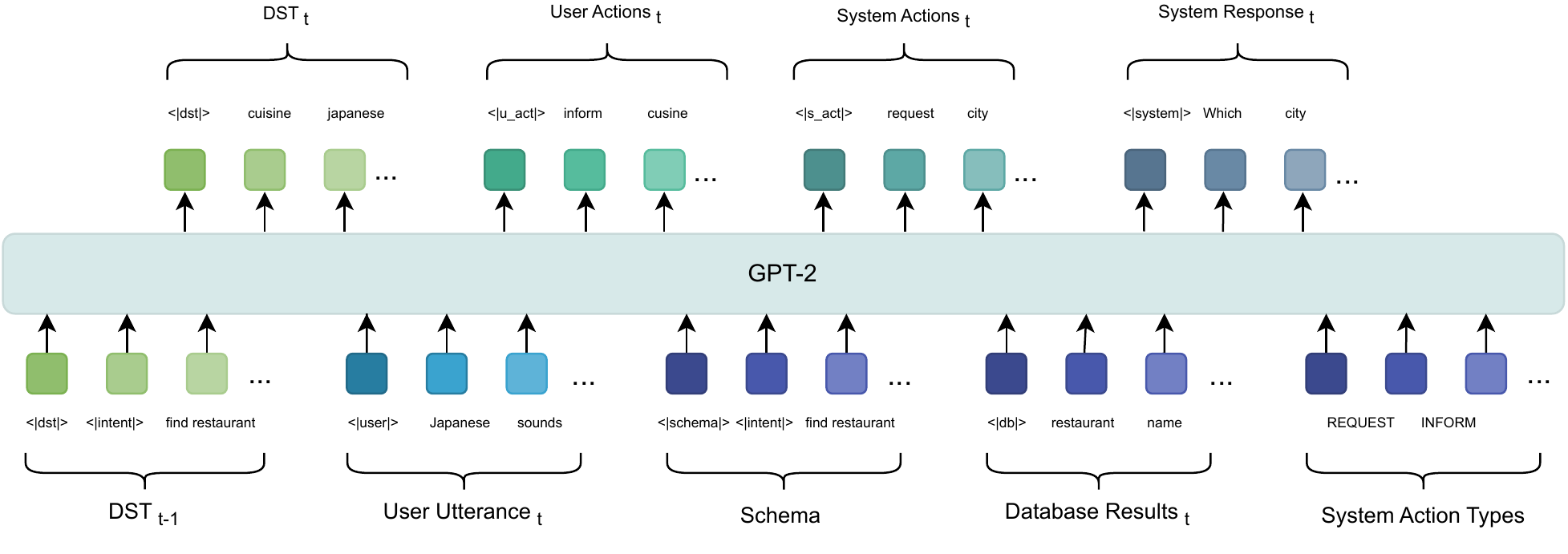}
   \vspace{-15pt}
   \caption{
       Overview of our approach. A GPT-2 model is fed the dialog state of the previous turn, the last user utterance, relevant schemas, database search results, and a list of system action names.
       As output, the model autoregressively generates the current dialog state, user actions, system actions, and system response.
   }
   \vspace{-15pt}
   \label{fig:our_model}
\end{figure*}

The state-of-the-art (SOTA) approaches in ToD systems formulate the problem as a conditional sequence generation task and finetune pre-trained causal language models in a supervised manner using single-domain or multi-domain datasets~\cite{HosseiniAsl2020ASL,Peng2021SoloistBT,Lee2020SUMBTLaRLEN,Yang2020UBARTF,Jeon2021DORATP,Sun2022BORTBA,Yang2022UBARv2TM}.
These systems feed the dialog history to the model as input and the output is a cascaded generation~\cite{su2021multi} of the DST, POL, and NLG.
%
%
Such systems are typically trained using large amounts of labeled data. 
In order for a dialog system to perform well on a specific task or in a specific domain, it needs to be trained on a large amount of labeled data that is specific to that task or domain.
A major drawback of most of these systems is that they fail to generalize to unseen domains and acquiring data for each new domain is prohibitively laborious and expensive, which motivates zero-shot generalizable ToD systems.
Recently, researchers have shown the possibility of building zero-shot generalizable individual components like the DST, next action prediction, and response generation, for ToD systems~\cite{Lee2021DialogueST,siddique2021linguistically,Mehri2021SchemaGuidedPF,siddique2021generalized}.
Nonetheless, to the best of our knowledge there has been no work on building zero-shot generalizable end-to-end ToD systems.

We propose a novel \textbf{Z}ero-\textbf{S}hot generalizable end-to-end \textbf{ToD} system, {\oursys}\footnote{The code is available at \url{https://github.com/MultifacetedNLP/ZS-ToD}}, using context summarization and domain schema.
The domain information can be represented in the form of a schema, which contains a set of intents and the 
relevant set of slots needed to fulfill a given intent.
The domain schema can facilitate zero-shot generalization to unseen domains in ToD systems. 
Figure~\ref{fig:approach} shows
a few example turns from a dialog, the intermediate outputs consisting of the DST, user actions, system actions, and the domain schema for that dialog.
In the first turn, the schema guides the system to infer the user intent, and the first few system actions are based on the required slots of the schema.
As the conversation continues, the user makes additional queries that are part of the optional slots. 
Similarly, when the system queries the database using the query parameters from the DST, the results contain the slot names listed in the result slots. 
The top search result is usually offered to the user and if the user rejects
the offered item, the next option from the search results is offered.

SOTA ToD systems use dialog history as the context to generate a response for a given turn. 
Generally, the dialog history consist of multiple turns (e.g., more than 20 turns) often containing conflicting information. For example, a slot value can be updated in later turns or the active intent of the user could change. 
At each turn, the system needs to model long as well as short range dependencies in the dialog context to accurately predict the current dialog state, and any errors made at any step would propagate to future steps. 
To alleviate this problem, we propose to replace the dialog history with the
dialog state from the previous turn, as this would provide a concise summary of all the previous turns and allow the system to focus on the current state rather than previous states.
Using a summarized context reduces the context size by a significant amount, thus allowing us to feed extra information, such as domain schema, database results and list of system action types, which otherwise would not have been possible without using a larger language model.

The goal of ToD systems is to generate a response (and intermediate outputs if needed), so there should be an explicit focus on the loss calculation for the response.
To fulfill this requirement, we propose a two-step training process, where the first step focuses on understanding the structure of the dialog data, and the second step focuses on generating the correct response. 
SOTA {\tod} models are passed an input prompt (i.e, dialog history) and generate a response for the given prompt.
However, these systems are trained using a {\celoss} over the whole sequence, i.e, dialog history and response. 
This kind of loss calculation for optimization might give the model superfluous rewards for correctly predicting the input prompt.


To evaluate the effectiveness of our proposed model, we conducted extensive evaluations using Schema Guided Dialogue (SGD) and SGD-X dataset that span multi-domain dialogs across 20 domains.
{\oursys} outperforms existing baseline systems across key metrics, particularly with a \textbf{+17\% joint goal accuracy} and \textbf{+5 inform} improvement over prior work, demonstrating the feasibility of our approach for {\zs} generalizable {\etoe} {\tod} systems.


\section{Methodology}

\subsection{Pre-trained Language Models}
Language models (e.g., BERT~\cite{Devlin2019BERTPO}, GPT-2~\cite{Radford2019LanguageMA}) have been trained on massive amounts of textual data and have shown state-of-the-art results in a variety of NLP tasks.
In this work, we use GPT-2 as the base model and fine-tune it on task-oriented dialog data to build a zero-shot generalizable end-to-end ToD system.
The GPT-2 model is pre-trained for autoregressive generation (i.e., predicting the next token given past tokens) on the WebText dataset (i.e., 40 GB of textual data) and adapts a transformer-based neural architecture~\cite{vaswani2017attention}.
Suppose we have a natural language sequence  $(\smalls_1, \cdots , \smalls_\smalln)$ where symbol $\smalls_\smalli$ is drawn from a fixed set of symbols. The sequential ordering of language leads to factorizing the joint probabilities over symbols as a product of conditional probabilities~\cite{bengio2003neural}, as given below.
\vspace{-5pt}
$$
p(\smalls) = \prod_{i=1}^\smalln p(\smalls_\smalli | \smalls_1, \cdots ,\smalls_{\smalli-1})
\vspace{-5pt}
$$

Using this approach, it is possible to estimate $p(\smalls)$ and any conditionals of the form 
$p(\smalls_{\smalli - k}, \cdots , \smalls_\smalli | \smalls_1, \cdots , \smalls_{\smalli - k - 1} )$, and perform tractable sampling. 
Since we formulate our problem as a sequence generation problem, GPT-2 is a natural choice for our TOD system.



\subsection{Problem Formulation}
We formulate task-oriented dialog response generation as a conditional sequence generation task. 
To facilitate zero-shot generalization to unseen domains, we condition the response generation for a dialog $x$ on the domain schema $S_i$, in addition to the dialog context.
The overall loss of our model can be written as:

\vspace{-10pt}
\begin{equation}    
    \mathcal{L}_{ZS-ToD} = \mathbb{E}_{x \sim D} \biggl( \sum_{n=1}^{T} -\log p(x_n | x_{<n}, S_i) \biggl)
    \label{eq:schema_gen}
\end{equation}

Specifically, in a multi-domain dialog system, the domain semantics are encapsulated in the domain schema denoted by $S = \{S_1, S_2, \cdots \}$ where $S_i$ represents schema for domain~$D_i$, which consists of a set of intents $I = \{I_1, I_2, \cdots \}$ and relevant set of slots $K = \{K_1, K_2, \cdots \}$ to fulfill a given intent.
A dialog session is composed of interactions between the user and the system in natural language from (single or) multiple domains.
We denote the dialog as 
$\{U^u_1, U^s_1, \cdots, U^s_T\}$ where $U^u_i$ and $U^s_i$ represents user and system utterances, respectively, at turn $i$ and $T$ is the number of turns in a dialog. 
We summarize the dialog history up to turn $t-1$ in the form of a dialog state represented by $DS_{t-1}$, which tracks the user's active intent $I_k$ and a list of tuples recording the slot names and corresponding slot values in a particular domain $(D_j, K_i, V_i)$, where $V_i$ represents the value for the slot $K_i$.
 At turn $t$, {\oursys} estimates the probability of the dialog state $DS_t$ by conditioning on previous dialog state $DS_{t-1}$, user's current utterance $U^u_t$,  and domain schema $S_i$:
\vspace{-1pt}
\begin{equation}    
    P(DS_t | DS_{t-1}, U^u_t, S_i)
    \label{eq:dialog_state}
\end{equation}





Then, at turn $t$, {\oursys} estimates the probability of the user action $A^u_t$ by conditioning on dialog state $DS_t$, user's current utterance $U^u_t$, and domain schema $S_i$:
\vspace{-1pt}
\begin{equation}
    P(A^u_t | DS_{t}, U^u_t, S_i )
    \label{eq:user_action}
\end{equation}
where the user action $A^u_t$ is represented by a list of tuples $(D_j, a^u_n, K_i, V_i)$ to record the user's action type $a^u_i \in \{a^u_1, a^u_2, \cdots, a^u_m \}$, slot name $K_i$, and corresponding slot values $V_i$ in a particular domain $D_j$.
The estimated dialog state $DS_t$ at turn $t$ is used to query the database (if needed), which returns a list of database results denoted by $DB_t$, that satisfy the constraints in the dialog state.

Next, {\oursys} estimates the probability of the system action $A^s_t$ denoted by a list of tuples $(D_j, a^s_m, K_i, V_i)$ where $a^s_i \in \{a^s_1, a^s_2, \cdots, a^s_n \}$ represents system's action type by conditioning on the dialog state $DS_{t}$, user utterance $U^u_t$, user action $A^u_t$, database results $DB_t$, the set of all available system action types $\forall a^s_i$:
\vspace{-1pt}
\begin{equation}
    P(A^s_t | DS_{t}, U^u_t, A^u_t, DB_t, \forall a^s_i)
    \label{eq:system_action}
\end{equation}

Finally, {\oursys} estimates the probability of the system's natural language response $U^s_t$ by conditioning on the dialog state $DS_t$, user utterance $U^u_t$, user action $A^u_t$, system action $A^s_t$, and domain schema $S_i$:
\vspace{-1pt}
\begin{equation}
    P(U^s_t | DS_t, U^u_t,  A^u_t, A^s_t, S_i)
    \label{eq:system_response}
\end{equation}

To accommodate for multi-domain dialogs, multiple domain schemas can be used to condition on.
In the traditional supervised learning setting, we have labeled training dialogs from all the domains. 
Whereas, in the zero-shot learning setup we assume that the training dialogs are available only for seen domains $D_s = \{D_1, D_2, \cdots, D_k \}$ and the dialogs from unseen domains $D_u = \{ D_{k+1}, D_{k+2}, \cdots \}$ may only show up at inference time where $ D_s \cap D_u =  \emptyset$. 
This challenging setting is the focus of the paper.  





\subsection{Model Architecture}
We use the pre-trained GPT-2 with a language modeling (LM) head.
Following the autoregressive nature of the model and problem formulation, we feed the previous dialog state, the user's current utterance, domain schema, and optionally database results to {\oursys} after tokenization. 
Then, {\oursys} outputs a representation, which upon decoding represents the updated dialog state, user actions, system actions, and system response in natural language. In this work, we employ the greedy decoding strategy for text generation. 

\begin{table*}[!t]
    \centering
    \begin{adjustbox}{max width=\textwidth}
        \begin{tabular}{|c|c|c|c|c|c|c|c|c|c|c|c|c|c|c|}
            \hline
            \multirow{3}{*}{Model}     &         & \multirow{2}{*}{\textbf{Intent}}   & \textbf{Requested} & \textbf{Average}        & \textbf{Joint}          &                &         & \textbf{Average}  & \textbf{Joint}    & \textbf{Average}    & \textbf{Joint}      & \multirow{2}{*}{\textbf{Response}} &          \\
                                       & \textbf{Domains} & \multirow{2}{*}{\textbf{Accuracy}} & \textbf{Slots}     & \textbf{Goal}           & \textbf{Goal}           & \textbf{Inform}         & \textbf{Success} & \textbf{Action}   & \textbf{Action}   & \textbf{UserAction} & \textbf{UserAction} & \multirow{2}{*}{\textbf{GLEU}}     & \textbf{Combined} \\
                                       &         &                           & \textbf{F1}        & \textbf{Accurracy}      & \textbf{Accuracy}       &                &         & \textbf{Accuracy} & \textbf{Accuracy} & \textbf{Accuracy}   & \textbf{Accuracy}   &                           &          \\ \hline
            \multirow{3}{*}{SimpleTOD} & all     & 78.60                     & 94.08     & 47.85          & 24.18          & 55.65          & 47.27   & 49.08    & 37.66    & 66.42      & 57.46      & 20.64                     & 72.10    \\
                                       & seen    & 80.07                     & 94.55     & 52.00          & 29.35          & 58.35          & 50.13   & 51.43    & 40.26    & 68.88      & 60.31      & 24.89                     & 79.13    \\
                                       & unseen  & 78.63                     & 93.92     & 46.27          & 22.72          & 54.28          & 46.17   & 48.29    & 37.12    & 65.55      & 56.65      & 19.24                     & 69.47    \\ \hline
            {SimpleTOD w/}             & all     & 82.34                     & 95.72     & 58.03          & 30.36          & 68.30          & 60.47   & 55.18    & 43.42    & 70.30      & 60.23      & 22.03                     & 86.41    \\
            {Schema \&}                & seen    & 83.32                     & 96.05     & 61.29          & 34.88          & 70.05          & 62.68   & 57.28    & 46.01    & 72.34      & 62.61      & 25.68                     & 92.04    \\
            {DB Results}               & unseen  & 82.19                     & 95.71     & 57.35          & 29.20          & 68.10          & 60.48   & 54.64    & 42.85    & 70.19      & 60.24      & 20.40                     & 84.69    \\ \hline
            \multirow{3}{*}{\textbf{\oursys}}   & all     & \textbf{84.83}                     & \textbf{95.53}     & \textbf{72.38} & \textbf{48.44} & \textbf{73.08} & \textbf{62.19}   & \textbf{58.32}    & \textbf{46.31}    & \textbf{73.20}      & \textbf{64.20}      & \textbf{20.04}                     & \textbf{87.67}    \\
                                       & seen    & \textbf{85.48}                     & \textbf{95.88}     & \textbf{74.23} & \textbf{52.05} & \textbf{74.72} & \textbf{63.85}   & \textbf{60.19}    & \textbf{48.69}    & \textbf{74.89}      & \textbf{66.24}      & \textbf{24.66}                     & \textbf{93.95}    \\
                                       & unseen  & \textbf{84.45}                     & \textbf{95.42}     & \textbf{72.03} & \textbf{47.83} & \textbf{71.68} & \textbf{61.63}   & \textbf{57.42}    & \textbf{45.21}    & \textbf{72.56}      & \textbf{63.46}      & \textbf{18.51}                     & \textbf{85.16}    \\ \hline
        \end{tabular}
    \end{adjustbox}
    \vspace{-5pt}
    \caption{Main Results. For end-to-end systems,~\oursys~outperforms existing baselines across all metrics, particularly there is significant improvement in key metrics like Average/Joint Goal Accuracy and Inform.}
    \label{tab:main-results}
    \vspace{-15pt}
\end{table*}

\subsection{Two-step Training}
Following the conditional generation in GPT-2, at a given turn $t$, 
we pass input tokens $S_{in} = \{s_1, \cdots, s_k\}$ as the prompt and the model generates a response $S_{out} = \{s_{k+1}, \cdots, s_{m+k} \}$, where $k$ and $m$ represent the input and output lengths, respectively.
Since conditional generation models (e.g., GPT-2) process one token at a time, the probability of predicting a token $\smalls_\smalli$ can be written as:
$p(\smalls_\smalli | \smalls_1, \cdots ,\smalls_{\smalli-1})
$.
The standard procedure for training these models is to optimize the Cross-Entropy (CE) loss over the full sequence. 
In ToD systems,
the input prompt is usually a long sequence of text that contains the entire dialog history, and the generation is much shorter than the input prompt.
Since the focus is not on generating the input prompt, we need to modify the loss function to pay less attention to the input prompt and more attention to the response.

To overcome the aforementioned issue, we propose a two step training approach.
In the first step, we follow the standard training procedure and calculate the CE loss on the full sequence.
For the second step, we initialize the model with the weights from the first step and calculate the CE loss only on the response,
as shown in Equation~\eqref{eq:loss_func}.
\vspace{-5pt}
\begin{equation}
    L = - \sum_{i=k+1}^{k+m} s_i \log(p_i)
    \label{eq:loss_func}
\end{equation}
\vspace{-10pt}

\subsection{Zero-shot Generalization}
Once {\oursys} is trained using the above-mentioned techniques, it can generalize to unseen domains seamlessly.
When {\oursys} is exposed to dialogs from a new unseen domain, the domain schema is expected to be part of the input.
Since the problem is formulated as a conditional generation, {\oursys} pays attention to the relevant parts of the schema to generate the user intent, slot names as well as slot values, thus adapting to the new unseen domains with no additional training.

\section{Experimental Setup}

\subsection{Datasets}

\stitle{The Schema Guided Dialogue (SGD).} SGD dataset is a large-scale dataset for task-oriented dialogue that consists of over 16K multi-domain dialogs between a human and a virtual assistant covering 20 domains. The dataset also provides a schema for each domain that
provides a textual description of the domain, a list of slots, and a list of intents. A slot contains a name, textual description,
and possible values for categorical slots and an intent contains a name, textual description, optional slots, and result slots.

\stitle{SGD-X}. SGD-X dataset is an extension of the SGD dataset that contains stylistic variants for every schema in SGD.
It provides 5 variants of domain schemas, where each variant incrementally moves further away from the original schema.
The goal of this dataset is to evaluate model sensitivity to schema variations.
The authors of the dataset have shown that two of the top-performing schema-guided DST models are sensitive to schema changes and have had significant performance drops on SGD-X.

\subsection{Evaluation Metrics}

To evaluate the performance of our model, we compute multiple metrics on each component of a ToD system.

\stitle{DST.} We evaluate the performance of DST by calculating the intent accuracy, average goal accuracy (AGA), joint goal accuracy (JGA), and requested slot F1.

\stitle{System Actions.} To evaluate the system actions, we compute the following metrics: inform, success, average action accuracy (AAA), and joint action accuracy (JAA).
Inform measures whether a system has provided a correct entity and success measures whether it has answered all the requested
information. AAA and JAA are similar to the goal metrics and are calculated from system actions. 
For inform, from the ground truth system actions we filter actions by action type inform (Inform, Inform Count)
and check if they are predicted correctly. For success, we filter actions by slot names that are in the requested slots and
check if the action slot values are predicted correctly. AAA and JAA are implemented following the implementations of AGA and JGA.
Since we also predict user actions, we calculate the
average and joint accuracy of the predicted user actions.

\stitle{System Response.} For evaluating the system response, we report the GLEU~\cite{wu2016googles} score as it performs better on individual sentence pairs.

\stitle{Overall.} To get an overall score for the model, we calculate the combined score~\cite{mehri2019structured}: (Inform + Success) $\times$ 0.5 + GLEU.

To ensure a fair comparison of {\oursys} with existing systems that have reported results on the SGD dataset,
we use the evaluation script provided by the SGD dataset, where applicable.

\begin{table*}[!t]
    \centering
    \begin{adjustbox}{max width=\textwidth}
        \begin{tabular}{|c|c|c|c|c|c|c|c|c|c|c|c|c|}
            \hline
            \multirow{3}{*}{Model}    &         & \multirow{2}{*}{Intent}   & Requested & Average        & Joint          &                &         & Average  & Joint    & \multirow{2}{*}{Response} &          \\
                                      & Domains & \multirow{2}{*}{Accuracy} & Slots     & Goal           & Goal           & Inform         & Success & Action   & Action   & \multirow{2}{*}{GLEU}     & Combined \\
                                      &         &                           & F1        & Accurracy      & Accuracy       &                &         & Accuracy & Accuracy &                           &          \\ \hline
            \multirow{2}{*}{\textbf{\oursys}} & all     & \textbf{84.83}                     & \textbf{95.53}     & \textbf{72.38} & \textbf{48.44} & \textbf{73.08} & \textbf{62.19}   & \textbf{58.32}    & \textbf{46.31}    & \textbf{20.04}                     & \textbf{87.67}    \\
            \multirow{2}{*}{\textbf{(this work)}}                          & seen    & \textbf{85.48}                     & \textbf{95.88}     & \textbf{74.23} & \textbf{52.05} & \textbf{74.72} & \textbf{63.85}   & \textbf{60.19}    & \textbf{48.69}    & \textbf{24.66}                     & \textbf{93.95}    \\
                                      & unseen  & \textbf{84.45}                     & \textbf{95.42}     & \textbf{72.03} & \textbf{47.83} & \textbf{71.68} & \textbf{61.63}   & \textbf{57.42}    & \textbf{45.21}    & \textbf{18.51}                     & \textbf{85.16}    \\ \hline
            {w/o}                     & all     & 75.08                     & 92.80     & 62.47          & 39.52          & 48.13          & 44.27   & 40.38    & 30.71    & 11.41                     & 57.61    \\
            {Two Step}                & seen    & 75.75                     & 93.13     & 64.66          & 42.76          & 50.26          & 46.47   & 41.96    & 32.42    & 13.75                     & 62.11    \\
            {Training}                & unseen  & 75.79                     & 92.90     & 62.60          & 39.25          & 47.55          & 44.32   & 40.25    & 30.66    & 11.03                     & 56.97    \\ \hline
            {w/o}                     & all     & 82.14                     & 94.67     & 64.70          & 38.47          & 59.88          & 53.88   & 54.14    & 43.07    & 21.15                     & 78.03    \\
            {Domain}                  & seen    & 83.34                     & 95.10     & 67.62          & 43.39          & 62.30          & 56.64   & 56.61    & 45.92    & 27.10                     & 86.57    \\
            {Schema}                  & unseen  & 81.96                     & 94.52     & 63.95          & 37.59          & 58.65          & 53.25   & 53.22    & 42.20    & 19.33                     & 75.28    \\ \hline
            {w/o}                     & all     & 82.50                     & 95.48     & 71.54          & 43.20          & 50.96          & 56.89   & 53.67    & 41.73    & 17.62                     & 71.54    \\
            {DB}                      & seen    & 83.26                     & 95.85     & 73.87          & 47.62          & 53.03          & 59.08   & 55.73    & 43.91    & 23.12                     & 79.17    \\
            {Results}                 & unseen  & 82.19                     & 95.36     & 71.04          & 42.17          & 50.33          & 56.95   & 53.17    & 41.52    & 16.07                     & 69.70    \\ \hline
            {w/o Sys}                 & all     & 82.56                     & 96.00     & 72.86          & 44.52          & 60.13          & 61.91   & 57.98    & 45.86    & 21.02                     & 82.04    \\
            {Action}                  & seen    & 83.25                     & 96.32     & 75.11          & 48.77          & 61.69          & 64.04   & 60.12    & 48.37    & 26.56                     & 89.43    \\
            {Names}                   & unseen  & 82.38                     & 95.91     & 72.44          & 43.60          & 59.61          & 61.75   & 57.29    & 45.26    & 19.16                     & 79.84    \\ \hline
        \end{tabular}
    \end{adjustbox}
    \caption{Ablation Study of~\oursys.}
    \label{tab:ablation-results}
    \vspace{-15pt}
\end{table*}

\section{Results}
\begin{table}
    \begin{adjustbox}{max width=0.45\textwidth}
        \begin{tabular}{|c|c|c|c|c|}
            \hline
            \multirow{2}{*}{Model} & Intent   & Requested & Average & Joint \\
                                   & Accuracy & Slot F1   & GA      & GA    \\ \hline
            SGD Baseline           & 90.60    & \textbf{96.50}     & 56      & 25.40 \\ \hline
            FastSGT                & 90.33    & 96.33     & 60.66   & 29.20 \\ \hline
            Seq2Seq-DU             & \textbf{91.00}    & -         & -       & 30.10 \\ \hline
            DSGFNET                & -        & -         & -       & 32.10 \\ \hline
            \textbf{\oursys}              & 81.49    & 95.97     & \textbf{74.08}   & \textbf{49.73} \\ \hline 
            
        \end{tabular}
    \end{adjustbox}
    \vspace{-5pt}
    \caption{Results on SGD test set. Our approach significantly outperforms baselines methods in terms of average and joint goal accuracy.}
    \label{tab:other-results}
    \vspace{-15pt}
\end{table}

\stitle{Main Results.}
Since no E2E ToD system has reported results for the SGD dataset, we follow~\cite{HosseiniAsl2020ASL} to implement some of the popular baseline methods to compare with our approach and present the results in Table~\ref{tab:main-results}.
We can see that~\oursys~outperforms all the baselines across all metrics except GLEU, where its performance is super competitive (e.g., 24.89 vs 24.66). 
An explanation of this could be that since we replaced the dialog history with the dialog state, the performance of the model improved on all other metrics, but the model lost a lot of exposure to dialog utterances.
Another reason could be greedy decoding which works well for a structured generation but is not the best strategy for fluent text generation.
While the system response requires a fluent generation, all other parts of the generation can be deemed as a structured generation.
On the other hand, nucleus and top-k sampling strategies are better suited for a fluent generation but are not the best for a structured generation.
We formulated the problem as a single sequence generation, and we can only select one strategy, so there is bound to be a trade-off. 
Since there is no single strategy that is best suited for both fluent and structured generation, our selection of greedy decoding may have been the cause for the loss of fluency in response generation.

We evaluate the DST performance of~\oursys~with the evaluation script provided by the SGD dataset and present our results alongside
other baseline DST models in Table~\ref{tab:other-results}.
We can see that even though our method is not specifically designed for DST, still it significantly outperforms the baselines models in the
important metrics: Average and Joint Goal Accuracy.

\begin{figure}[!t]
    \centering
    \includegraphics[width=0.9\linewidth]{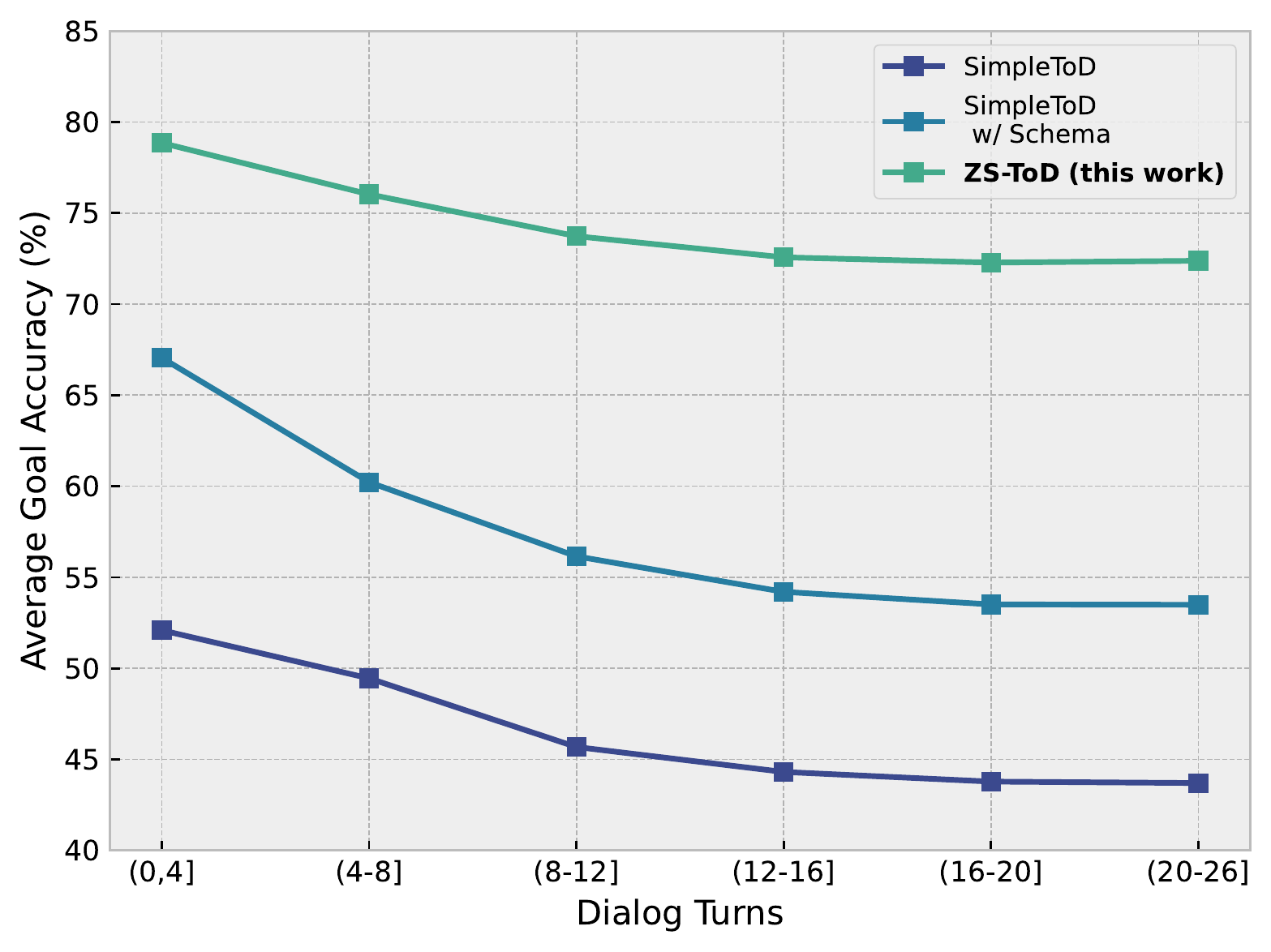}
    \vspace{-10pt}
    \caption{
        Performance of dialog systems on the SGD test set with respect to dialog turns
    }
    \label{fig:dialog_turns}
    \vspace{-15pt}
\end{figure}

\stitle{Long Range Dialog Dependencies.}
In order to process dialogs that have a large number of turns, a system must be effective at capturing long-range dependencies.
To test this ability, we group the test dialogs based on the number of turns and evaluate on each group.
As shown in Figure~\ref{fig:dialog_turns},~\oursys~outperforms the baseline systems across all groups.
Generally, in the first few turns of a dialog, the main focus is on figuring out what the user wants. The user could switch among mulitple options before finally deciding on one, however
towards the end of a dialog, usually the user has a clear idea of what he or she wants, so is less likely to make many changes.
For the first few turns, we have observed that there is a steeper drop in performance of the baseline when compared to~\oursys~.
A possible explanation of this could be that, since we pass the dialog summary to the model, it contains the correct state of the dialog at the previous turn, which helps the model to make better predictions.
In groups with large number of turns, both the baseline and~\oursys~perform similarly, which suggests even though~\oursys~does well in capturing
medium range dependencies, long range dependencies are still a challenge.

\begin{figure}[t]
    \centering
    \includegraphics[width=0.9\linewidth]{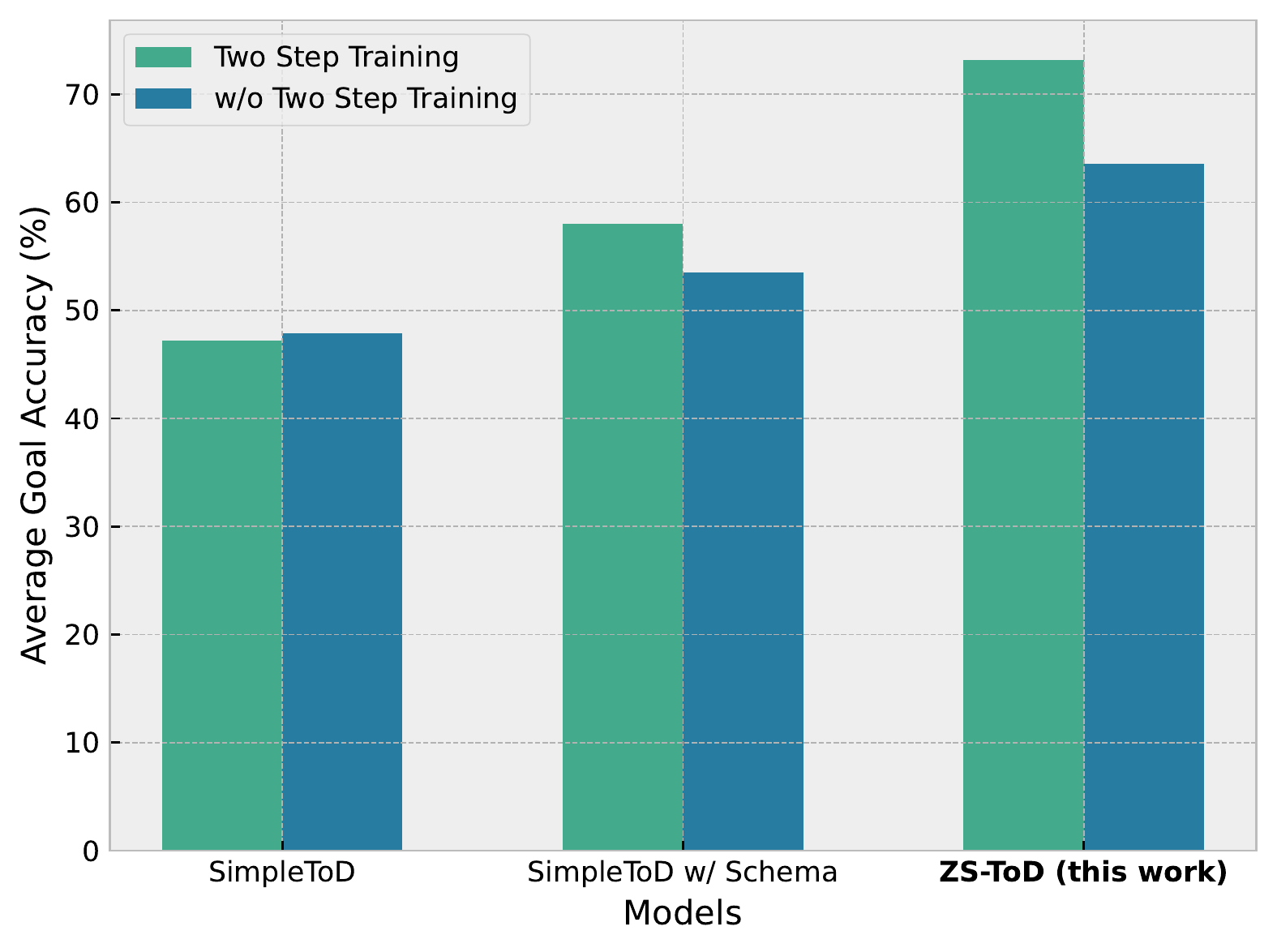}
    \vspace{-10pt}
    \caption{
        Effect of Two Step Training on dialog systems
    }
    \vspace{-15pt}
    \label{fig:two_step_training}
\end{figure}

\stitle{Two Step Training.}
To better understand the effect of the two step training process, we compared {\oursys} and a few baseline systems with and without the two step training process.
In Figure~\ref{fig:two_step_training}, we can see that models that incorporate schema benefit from the two step training process.

\stitle{Ablation Study.}
To get a better understanding of the different components of our model, we drop a certain component of {\oursys} to show the effect on the performance
and report an ablation study in Table~\ref{tab:ablation-results}.
We can see that dropping two step training drastically degrades performance across all metrics, which suggests the importance of the training mechanism for {\oursys}.

The role of schema is also important as we can see that the performance of {\oursys} drops across all metrics when we drop schema.
Another important aspect to notice here is that this variant has the largest difference in performance between seen and unseen domains.
These observations indicate that schema not only aids the model to generalize to new domains, but also plays a central role in the overall performance of the system.
When the database results are excluded from the input, there is a big drop in 
metrics related to system actions. Additionally, there is a small drop in the DST performance as well, which suggests that
there is some correlation between the database results and DST.
When we omit the list of system actions types, the metrics related to system actions decreases the most, particularly Inform, but the drop in performance is much less when compared to the setting when the database results were dropped. However, in this setting there were no changes to metrics related to DST.

\begin{figure}[!t]
    \centering
    \includegraphics[width=0.9\linewidth]{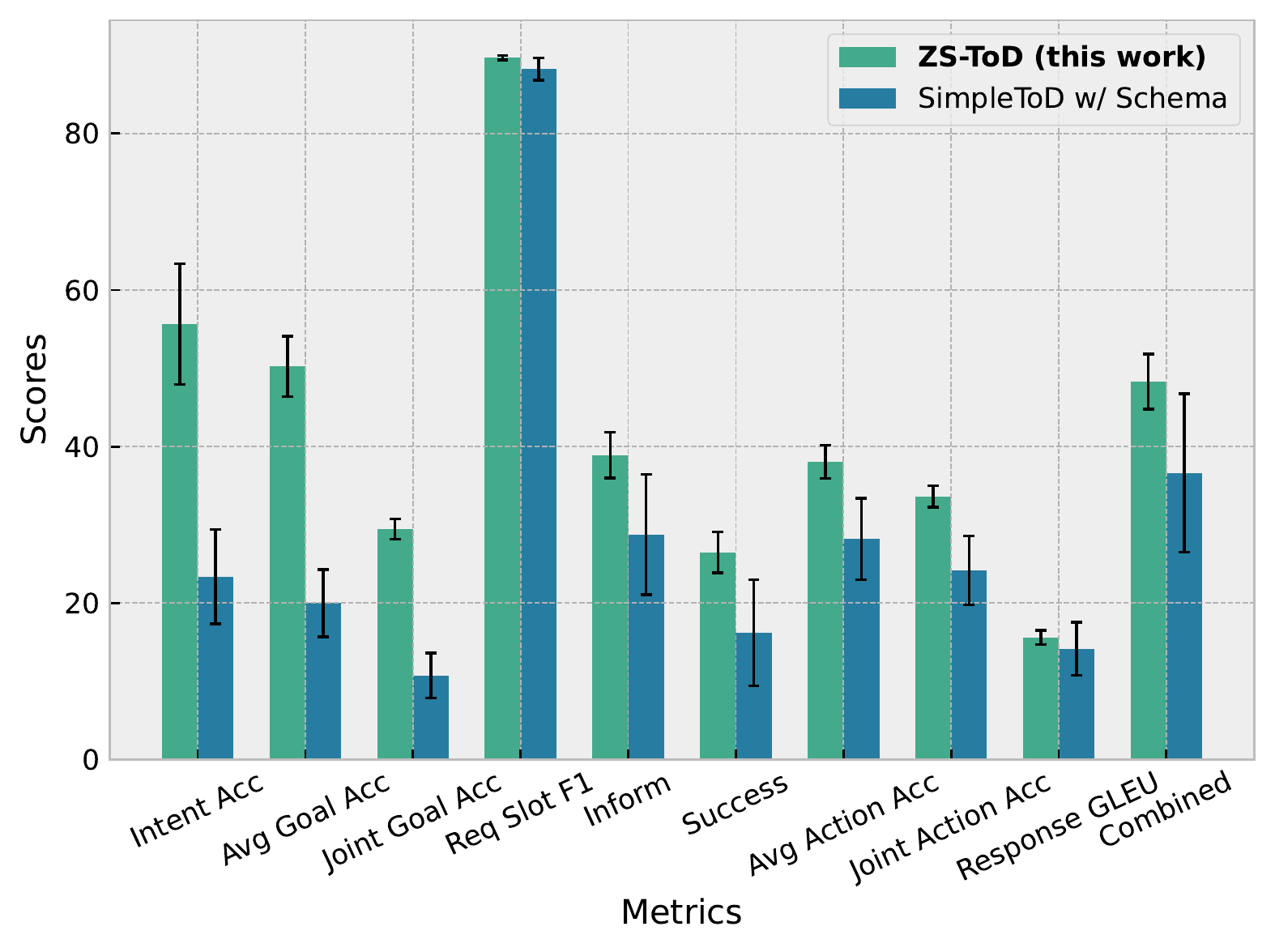}
    \vspace{-15pt}
    \caption{
        {\sgdx} results: Mean and standard deviation of each metric across all versions of {\sgdx}
    }
    \vspace{-15pt}
    \label{fig:sgdx_graph}
    
\end{figure}

\stitle{Results on {\sgdx}.}
To access the robustness of {\oursys}, we ran experiments on the unseen domains of the {\sgdx} dataset and present the results in Figure~\ref{fig:sgdx_graph}.
The bar graph shows the mean of each metric across all the versions of {\sgdx} and the error bars show the standard deviation.
{\oursys} outperforms the baseline across all metrics and has lower standard deviation, showing the robustness of {\oursys} to domain schema variations.

\section{Related Works}

\stitle{Supervised End to End Models.}
Pretrained language models like BERT~\cite{Devlin2019BERTPO}, GPT-2~\cite{Radford2019LanguageMA}, T5~\cite{Raffel2019ExploringTL} and UniLM~\cite{Dong2019UnifiedLM}
have been used extensively in the literature for {\etoe} models for {\tod} systems~\cite{HosseiniAsl2020ASL,Peng2021SoloistBT,Lee2020SUMBTLaRLEN,Yang2020UBARTF,Jeon2021DORATP,Sun2022BORTBA,Yang2022UBARv2TM,Noroozi2020AFA,He2021GALAXYAG}
on the popular MultiWoz 2.0~\cite{budzianowski2018large} dataset. Even though some of these models has shown experiments on zero shot performance,
they shine under supervised settings and are not able to generalize to new domains, whereas our model is designed to be zero shot generalizable. 
In all the existing systems, the dialog history is passed as the context, whereas we use a summarized context which consists
of the current user utterance and the DST of the previous state as the context.


\stitle{Schema Guided Models.} To obtain Zero Shot generalizability, some work has been done by incorporating schema to transfer knowledge across domains. However these systems
only focus on certain components of {\tod} systems, such as  DST~\cite{Feng2020ASA,Feng2022DynamicSG,Lee2021DialogueST,Noroozi2020AFA,Wang2022SlotDM}
and next action prediction and response generation~\cite{Mosig2020STARAS,Mehri2021SchemaGuidedPF}. 

\stitle{Description and Prompt Based Models.} Generally, schema is described using abbreviated notations or in snake case,  
and this vocabulary that is not usually present in natural language. To remedy this problem, there has been some work on description based DST~\cite{Zhao2022DescriptionDrivenTD,Lin2021LeveragingSD,Mi2021CINSCI}, where the abbreviated and unnatural words are converted into natural descriptions, from which models can obtain useful semantic descriptions. Another aspect of ToD systems
is slot filling, which has been formulated as a question answering problem~\cite{Yang2022PromptLF,Madotto2021FewShotBP,Hu2022InContextLF,Brown2020LanguageMA,Su2021MultiTaskPF,Lin2021ZeroShotDS,Li2021ZeroshotGI}, where a prompt is passed as a natural language question and the model predicts the slot value. These models are making the context larger and require large language models to fit such a big input, whereas we embark on the opposite direction and try to make the 
context smaller.

\section{Conclusion}
We have presented a novel schema-guided zero-shot generalizable end-to-end task-oriented dialog system that estimates a concise summary of the dialog history through the dialog state.
This system leverages domain schemas and effective dialog history summarization to allow for robust generalization to unseen domains, overcoming the major bottleneck of acquiring labeled training data for each new domain.
Additionally, a two-stage training methodology was introduced, wherein the model initially acquires an understanding of the general structure of the dialog data and subsequently optimizes the response generation process.
The experimental results show the superiority of our proposed {\oursys} over state-of-the-art models on key metrics, particularly with a \textbf{+17\% joint goal accuracy} and \textbf{+5 inform} improvement over prior work.
To better evaluate the effectiveness of the proposed components and training mechanisms, we have provided an ablation study that shows the significance of our contributions.



\balance 
\bibliography{custom}
\bibliographystyle{flairs}
\end{document}